\documentclass{article}

\usepackage[preprint]{corl_2025}

\usepackage[T1]{fontenc}
\usepackage{amsmath,amssymb,amsfonts}
\usepackage{booktabs}
\usepackage{graphicx}
\usepackage{multirow}
\usepackage{enumitem}
\usepackage{xcolor}
\usepackage{url}
\usepackage{hyperref}
\usepackage{siunitx}
\usepackage{microtype}
\usepackage{array}
\usepackage{makecell}
\usepackage{adjustbox} 

\hypersetup{
  colorlinks=true,
  linkcolor=blue,
  citecolor=blue,
  urlcolor=blue,
  pdftitle={VLX-Go: From Multimodal Alignment to Embodied Tracking through Multi-Stage Learning and On-Policy Refinement},
  pdfsubject={Technical report on vision-language short-horizon waypoint prediction}
}

\newcommand{\method}{VLX-Go}
\newcommand{\trackvla}{TrackVLA}
\newcommand{\evtbench}{EVT-Bench}
\newcommand{\stt}{STT}
\newcommand{\dtask}{DT}
\newcommand{\atask}{AT}

\newcommand{\img}{\mathbf{I}}
\newcommand{\feat}{\mathbf{f}}
\newcommand{\token}{\mathbf{z}}
\newcommand{\traj}{\boldsymbol{\tau}}

\newcommand{\vel}{\mathbf{v}}
\newcommand{\instr}{\mathbf{q}}

\title{\method{}: Robust Embodied Tracking through Multi-Stage Learning and On-Policy Refinement}

\author{
 \textbf{Kyusong Lee\textsuperscript{1}},
 \textbf{Heting Ying\textsuperscript{1}},
 \textbf{Tiancheng Zhao\textsuperscript{1,2}}
 \\
\\
 \textsuperscript{1}Om AI Research,
 \textsuperscript{2}Binjiang Institute of Zhejiang University
\\
 \small{
   \texttt{tianchez@zju-bj.com}
 }
}

\begin{document}
\maketitle

\begin{abstract}
Embodied visual tracking requires a robot to identify a designated target from visual observations and control its own motion to follow that target through an environment. Following an isolated person is a comparatively simple demonstration of embodied visual tracking, but following the intended person among distractors remains challenging. Existing methods exhibit a substantial drop in success from single-target to distractor tracking, motivating a focus on the shared spaces that person-following robots must navigate. We present \method{}, a vision--language waypoint policy built around the hypothesis that visual and language representations should be aligned before tracking-specific motion is learned. The policy combines complementary frozen visual encoders, multiscale temporal context, a transformer backbone, and a lightweight head for robot-centric waypoint prediction. Its training progresses through vision--language alignment, multimodal reasoning mid-training, and tracking specialization on 19,532 oracle episodes. Compared with 41.54\% distractor-tracking success for our earlier OmTrackVLA model, the complete multi-stage recipe achieves 55.80\%. We then explore on-policy refinement using filtered oracle labels for states visited by the policy. This preliminary study adds 2.42 percentage points of distractor-tracking success, reaching 58.22\%, while reducing collision rate by 1.99 pts. The final policy records the highest distractor- and ambiguous-tracking success rates among the compared methods. These results support staged multimodal learning as a promising direction for tracking among distractors and motivate further study of on-policy refinement for person following in shared spaces. Code is available at \url{https://github.com/om-ai-lab/VLX-Go}.
\end{abstract}


\section{Introduction}
\label{sec:introduction}

Embodied navigation requires a robot to translate visual observations and language into motion through its surroundings. Predicting a short sequence of robot-centric waypoints offers a direct interface between multimodal perception and local control. We study this interface in embodied visual tracking (EVT), where a robot identifies a language-specified person or object and follows it through a changing scene. The robot must preserve target identity through occlusion and distractors while maintaining a useful following distance. This setting is relevant to service robots, guidance, and human--robot collaboration, and provides the empirical focus of this report.

Following an isolated person is a comparatively simple setting: the robot has fewer competing candidates to distinguish. A successful single-person demonstration therefore leaves an important deployment challenge unresolved. In shared spaces, other people can enter the camera view, cross the target's path, or resemble the target. Distractor tracking (DT) tests this challenge directly. Benchmark results reported by TrackVLA~\cite{wang2025trackvla} show a substantial drop from single-target tracking (STT) to DT across methods: for example, Uni-NaVid falls from 25.70\% to 11.30\% success, while TrackVLA falls from 85.10\% to 57.60\%. The size of this gap varies, but it motivates our emphasis on improving DT as a practical requirement for person-following robots.

Traditional EVT systems typically decompose the problem into detection, association, tracking, and planning. Such pipelines are interpretable, but errors compound: a target-identity mistake corrupts the tracking state, which in turn misleads the planner. Recent vision--language--action (VLA) approaches instead learn a shared representation for visual recognition, language grounding, temporal reasoning, and action prediction. This direction builds on embodied language grounding~\cite{ichter2022saycan,driess2023palme}, direct vision--language action prediction~\cite{zitkovich2023rt2}, and open generalist robot policies~\cite{ghosh2024octo,kim2024openvla}. Uni-NaVid~\cite{zhang2025uninavid} studies a unified video-conditioned navigation formulation, while \trackvla{}~\cite{wang2025trackvla} specializes VLA learning for EVT and demonstrates strong performance on \evtbench{}. TrackVLA++~\cite{liu2025trackvlapp} further introduces explicit spatial reasoning and long-horizon target memory.

A short-horizon waypoint representation makes the connection between multimodal understanding and local control explicit. Waypoints express the intended motion in the robot's coordinate frame, while a low-level controller handles execution. For target following, the prediction must combine the language-specified identity with both the current spatial configuration and recent motion. This motivates a policy that brings visual history and language into a shared representation before decoding a local trajectory.

\method{} formulates local motion generation as vision--language-conditioned short-horizon waypoint prediction. It uses complementary frozen DINOv3~\cite{simeoni2025dinov3} and SigLIP~\cite{zhai2023siglip} encoders, a multiscale history representation, a pretrained transformer backbone~\cite{yang2025qwen3}, and a waypoint regression head. This design pairs a transformer backbone with an explicit spatial output: eight future robot-centric waypoints that can be passed to a low-level controller.

Our working hypothesis is that effective target following among distractors benefits from aligning visual features with language before learning tracking-specific motion. This follows the staged learning principle used by OmChat~\cite{zhao2024omchat} and LLaVA~\cite{liu2023llava}, which aligns visual features with a pretrained language backbone before visual instruction tuning. We adopt this principle for embodied tracking through multimodal alignment, multimodal reasoning mid-training, and tracking specialization. Stage~1 trains only the vision--language adapter. Stage~2 jointly trains the transformer backbone and adapter using both the Stage~1 alignment data and additional VQA data. Stage~3 then starts from the Stage~2 base checkpoint and applies SFT on tracking data. Finally, we fine-tune on on-policy data aggregated with DAgger~\cite{ross2011dagger}, filtering privileged oracle labels using target framing, proximity, and recent tracking history. The main benchmark results evaluate the resulting tracking model across all three tasks.

Starting from a baseline model~\cite{omtrackvla2025} performance 41.54\% DT success; the multi-stage policy improves the success rate to 55.80\%. This substantial observed gain motivates the staged recipe, while leaving the contribution of each stage to be isolated. We then investigate on-policy refinement with DAgger as a preliminary extension: oracle supervision on policy-visited states raises DT success by another 2.42 percentage points to 58.22\%. Its combination of higher DT success and fewer collisions is promising for further study of tracking in shared spaces.

In summary, this report contributes:
\begin{itemize}[leftmargin=*]
    \item \textbf{A vision--language waypoint policy.} \method{} combines frozen DINOv3 and SigLIP features, multiscale temporal context, a transformer backbone, and a lightweight decoder for short-horizon robot-centric waypoints.
    \item \textbf{Three-stage training followed by DAgger.} Adapter alignment, multimodal reasoning mid-training, and tracking SFT connect general multimodal learning to waypoint prediction. We describe the 19,532-episode oracle collection and its conversion into training examples, and a DAgger stage that filters privileged oracle labels.
    \item \textbf{Evaluation across tracking conditions.} \method{} reaches success rates of 58.22\% and 60.93\% on \dtask{} and \atask{}, the highest listed SR on these tasks. Results emphasize episode success and collision avoidance, with comparisons of the three-stage and DAgger policies across all three tasks. Interaction diagnostics provide additional context for further development.
\end{itemize}

\section{Related Work}
\label{sec:related}

EVT differs from passive video tracking because the tracking agent controls the camera and therefore influences future observations. Earlier approaches include active visual tracking with reinforcement learning~\cite{luo2018active}, modular visual-servoing systems, and methods combining visual foundation models with learned policies. Zhong et al.~\cite{zhong2024evt} combine visual foundation models with offline reinforcement learning to improve sample efficiency and generalization in embodied tracking. PoliFormer~\cite{zeng2025poliformer} scales transformer-based on-policy reinforcement learning for embodied navigation and can be adapted to tracking-oriented tasks.

\evtbench{}, introduced with \trackvla{}~\cite{wang2025trackvla}, evaluates language-conditioned embodied tracking in realistic indoor environments. Its \stt{}, \dtask{}, and \atask{} settings progressively stress persistent tracking, distractor rejection, and semantic target disambiguation. This task family exposes the tight coupling between target identification and control: uncertainty about whom to follow directly changes the robot trajectory.

Language-conditioned robot learning has progressed from using language models as high-level planners to directly predicting embodied actions. SayCan grounds language-model proposals in executable robot affordances~\cite{ichter2022saycan}, while PaLM-E~\cite{driess2023palme} injects continuous perceptual observations into an embodied multimodal language model. RT-2~\cite{zitkovich2023rt2} further established the vision--language--action formulation by co-fine-tuning vision--language models to emit robot actions.

Staged multimodal learning provides a foundation for adapting visual understanding to embodied tasks. OmChat~\cite{zhao2024omchat} combines projector alignment, progressive multimodal pretraining, and visual instruction tuning to support long-context image and video understanding. Its separation of alignment from subsequent multimodal learning motivates our alignment--mid-training--specialization sequence, in which general multimodal representations are ultimately optimized for tracking-specific waypoint prediction.

For embodied navigation, NaVid~\cite{zhang2024navid} and Uni-NaVid~\cite{zhang2025uninavid} use video-conditioned language models to predict navigation actions across tasks. \trackvla{}~\cite{wang2025trackvla} specializes the VLA formulation for EVT, jointly learning target recognition and action generation, while TrackVLA++~\cite{liu2025trackvlapp} adds explicit spatial reasoning and target-identification memory for occlusion and distractors. MiniCPM-V~\cite{yao2025minicpmv} explores multimodal models for visual understanding, providing complementary context for the use of smaller backbones. Our earlier OmTrackVLA~\cite{omtrackvla2025} provides an open-source policy that maps monocular video and language to short-horizon waypoints, together with model weights and a training pipeline. It serves as the direct predecessor and baseline for \method{}, which investigates staged multimodal training followed by oracle-supervised on-policy data aggregation with DAgger~\cite{ross2011dagger} for embodied tracking.

Fine-grained visual grounding provides complementary context for language-conditioned target identification. VLM-FO1~\cite{liu2025vlmfo1} combines complementary visual encoders with token-based region references, using region--language alignment followed by perception instruction tuning. VLX-Seek~1.5~\cite{omailab2026vlxseek15} continues this region-reference direction with broader embodied-scene training and explicit rejection of absent targets. These studies relate to our use of complementary visual representations and staged multimodal learning, while addressing region-level perception. \method{} instead connects visual history and language to robot-centric waypoint prediction and is evaluated through closed-loop tracking.

Embodied policies also benefit from advances in visual representation learning. Strong frozen visual representations are increasingly used to reduce the amount of task-specific perception training required by robot policies. CLIP~\cite{radford2021clip} established scalable language-aligned visual pretraining, while DINOv2~\cite{oquab2023dinov2} demonstrated strong task-agnostic dense features learned without supervision. SigLIP~\cite{zhai2023siglip} improves language--image alignment with a sigmoid objective, and DINOv3~\cite{simeoni2025dinov3} extends self-supervised visual representation learning at scale. \method{} combines the latter two families: DINOv3 provides dense spatial structure, while SigLIP supplies language-aligned semantics.

\section{Task Formulation}
\label{sec:task}

At time $t$, the robot receives a natural-language target description $\instr$ and a temporal window of monocular RGB observations,
\begin{equation}
\mathcal{I}_t = \{\img_{t-H+1},\ldots,\img_t\},
\end{equation}
where $H$ denotes the number of frames in the model history. The policy predicts $K$ future robot-centric waypoints,
\begin{equation}
\hat{\traj}_t =
\left\{(\hat{x}_{t,k}, \hat{y}_{t,k}, \hat{\theta}_{t,k})\right\}_{k=1}^{K}.
\end{equation}
A low-level controller converts the predicted local trajectory into robot motion.

Following \evtbench{}~\cite{wang2025trackvla}, we report success rate (SR) and collision rate (CR); higher SR and lower CR are better. Tracking rate (TR) describes adherence to the benchmark tracking conditions, but does not by itself establish overall model quality: useful obstacle detours, yielding, and target recovery can temporarily reduce it (Section~\ref{sec:discussion}). Because collision statistics depend strongly on implementation and episode dynamics, we additionally retain per-episode collision timing, human--robot distance, and recent forward velocity for diagnostic analysis.

\section{\method{}}
\label{sec:method}

\subsection{Architecture Overview}

\begin{figure*}[t]
    \centering
    \includegraphics[width=\textwidth]{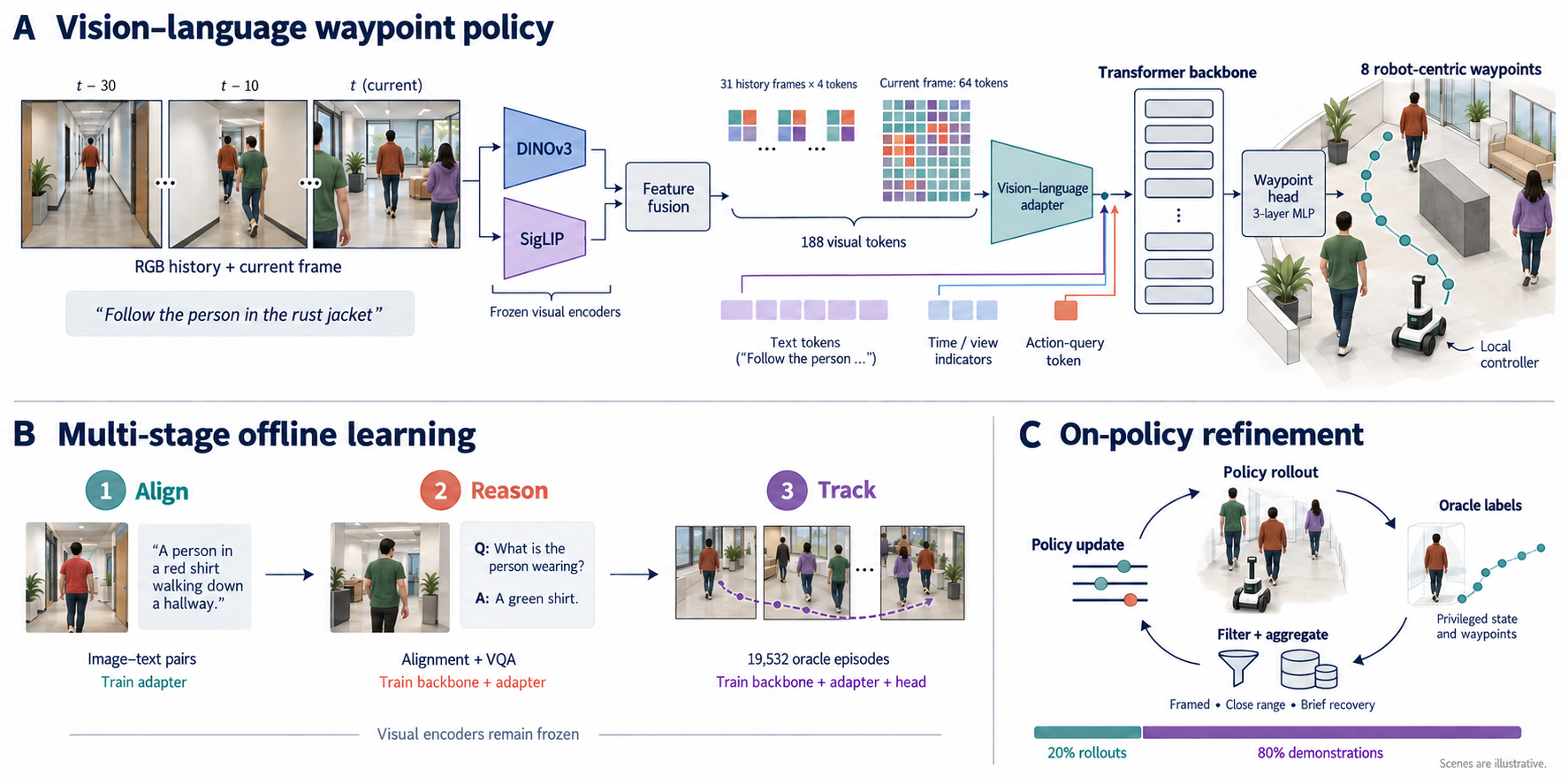}
    \caption{\method{} architecture and training workflow. \textbf{(A)} Frozen DINOv3 and SigLIP features are concatenated and pooled into 188 visual tokens: $31\times(2\times2)$ for history and $8\times8$ for the current frame. A vision--language adapter projects these features into a transformer backbone; temporal/view indicators, instruction tokens, and a learnable action query complete the input. A three-layer MLP decodes the action state into eight robot-centric waypoints. \textbf{(B)} Three offline stages train the adapter, then the backbone and adapter, and finally the backbone, adapter, and waypoint head. \textbf{(C)} DAgger collects policy rollouts, labels visited states with the oracle, filters the labels, and fine-tunes on aggregated rollouts and demonstrations. The updated policy supplies the next round of rollouts. Visual encoders remain frozen throughout. Scene and robot images are illustrative.}
    \label{fig:architecture}
\end{figure*}

Figure~\ref{fig:architecture} summarizes the architecture, offline training stages, and DAgger loop of \method{}. Historical and current RGB frames are encoded with frozen DINOv3 and SigLIP visual towers. The two feature maps are spatially aligned and concatenated. Historical frames are aggressively pooled into a small coarse grid, whereas the current frame retains a denser grid. These visual tokens are projected into the hidden space of a transformer backbone and concatenated with language tokens and a learnable action token. The hidden state of the action token is decoded into short-horizon robot-centric waypoints.

\subsection{Complementary Frozen Visual Encoders}

For each frame $\img_t$, we extract DINOv3 and SigLIP features,
\begin{align}
\feat^D_t &= E_D(\img_t),\\
\feat^S_t &= E_S(\img_t).
\end{align}
DINOv3 provides strong dense visual representations~\cite{simeoni2025dinov3}, whereas SigLIP provides image--language-aligned semantics~\cite{zhai2023siglip}. After spatial alignment, we concatenate the channels,
\begin{equation}
\feat_t =
\operatorname{Concat}\left[
\feat^D_t,\operatorname{Align}(\feat^S_t)
\right].
\end{equation}
In our default configuration, the DINOv3 branch contributes 384 dimensions and the SigLIP branch 1152, yielding a 1536-D concatenated feature before projection. Both visual towers are frozen. We precompute their outputs in FP16, decoupling expensive image encoding from policy optimization and enabling rapid controlled experiments.

\subsection{Multiscale Temporal Tokenization}

Dense visual tokens from every historical frame would make sequence length scale linearly with both history duration and spatial resolution. We therefore use asymmetric grid pooling. Historical frames are summarized with a coarse $G_c \times G_c$ grid, while the current frame uses a finer $G_f \times G_f$ grid:
\begin{equation}
\mathcal{Z}^{v}_t =
\left[
\mathcal{P}_{G_c}(\feat_{t-H+1}),\ldots,
\mathcal{P}_{G_c}(\feat_{t-1}),
\mathcal{P}_{G_f}(\feat_t)
\right].
\end{equation}
The default experiment uses $H=32$, $G_c=2$, and $G_f=8$. This preserves long temporal context at low token cost while allocating spatial detail to the observation that most directly determines the next control decision.

\subsection{Projection, Temporal Indicators, and Action Query}

A shared projector maps each 1536-D visual token to the transformer backbone's hidden dimension:
\begin{equation}
\tilde{\token}^{v}
=
W_2\,\sigma\!\left(
W_1\operatorname{LN}(\token^{v})+b_1
\right)+b_2,
\end{equation}
where $\sigma$ is GELU. Temporal/view indicators are added so that the transformer can distinguish frame age and coarse-history versus fine-current streams.

The natural-language instruction is tokenized using the backbone's tokenizer. We form the multimodal sequence
\begin{equation}
\mathcal{Z}_t =
\left[
\mathcal{Z}^{\mathrm{text}},
\mathcal{Z}^{\mathrm{history}},
\mathcal{Z}^{\mathrm{current}},
\token^{\mathrm{ACT}}
\right].
\end{equation}
The transformer backbone processes the sequence, and we read the final hidden state at the learnable action position,
\begin{equation}
\mathbf{h}^{\mathrm{ACT}}_t
=
\operatorname{Transformer}(\mathcal{Z}_t)[\mathrm{ACT}].
\end{equation}

\subsection{Waypoint Planner}

A small MLP maps $\mathbf{h}^{\mathrm{ACT}}_t$ to $3K$ values and reshapes them into $K$ robot-centric waypoints:
\begin{equation}
\hat{\traj}_t =
\operatorname{reshape}\left(
f_{\mathrm{wp}}(\mathbf{h}^{\mathrm{ACT}}_t), K, 3
\right).
\end{equation}
The default planner uses three layers. Given a validity mask $m_{t,k}$, the tracking objective is masked waypoint regression,
\begin{equation}
\mathcal{L}_{\mathrm{wp}} =
\frac{
\sum_{t,k}m_{t,k}
\left\|
\hat{\traj}_{t,k}-\traj_{t,k}
\right\|_2^2
}{
\sum_{t,k}m_{t,k}+\epsilon
}.
\end{equation}

\subsection{Three-Stage Training and DAgger}
\label{sec:training}

The model is trained sequentially through vision--language alignment, multimodal reasoning mid-training, and tracking SFT, and is then fine-tuned with on-policy data aggregation (DAgger). Table~\ref{tab:training_stages} summarizes the role and data of each step. Stage~2 continues from the aligned model, and its resulting base checkpoint initializes Stage~3.

\paragraph{Stage 1: adapter-only vision--language alignment.}
Only the vision--language adapter is optimized on the \texttt{blip\_laion\_cc\_sbu\_558k} image--text alignment collection and human recognition dataset; the visual encoders and transformer backbone remain frozen. This stage establishes the mapping between visual features and the language representation before updating the transformer backbone.

\paragraph{Stage 2: multimodal reasoning mid-training.}
Starting from Stage~1, we jointly train the transformer backbone and adapter using a mixture of the Stage~1 alignment data and general-purpose VQA data. This stage uses supervised text generation to develop multimodal representations before tracking specialization. All backbone parameters are updated. The output of this stage is the multimodal base checkpoint used for tracking specialization.

\paragraph{Stage 3: tracking supervised fine-tuning.}
We initialize from the Stage~2 base checkpoint and jointly train the backbone, adapter, and waypoint head on the oracle tracking collection described in Section~\ref{sec:data}, without adding other tracking datasets. The backbone is again updated through full-parameter SFT. We evaluate the resulting model on \stt{}, \dtask{}, and \atask{}. The results therefore measure the complete alignment--VQA--tracking training sequence. Both visual encoders remain frozen throughout training.

\paragraph{On-policy fine-tuning with DAgger.}
The three stages are trained offline on fixed datasets. We then fine-tune the Stage~3 model on a mixture of oracle demonstrations and filtered rollouts of the policy itself, labeled by the oracle (Section~\ref{sec:dagger}). Unlike Stages~1--3, this step requires closed-loop rollouts in the simulator. We call the Stage~3 model \method{} (three-stage) and the fine-tuned model \method{} (three-stage + DAgger).

\paragraph{Training objectives.}
Stages~1 and~2 use the standard autoregressive cross-entropy objective for supervised text generation. For a target sequence $y_{1:N}$ and multimodal context $c$,
\begin{equation}
\mathcal{L}_{\mathrm{text}} = -\frac{1}{N}\sum_{i=1}^{N}\log p_{\theta}(y_i\mid c,y_{<i}).
\end{equation}
The trainable parameters $\theta$ are restricted to the adapter in Stage~1 and include the adapter and full transformer backbone in Stage~2. Stage~3 and DAgger fine-tuning use the masked waypoint mean-squared error $\mathcal{L}_{\mathrm{wp}}$ defined above to optimize the tracking policy.

\begin{table}[t]
\centering
\small
\caption{Training workflow: three offline stages followed by on-policy DAgger fine-tuning (Section~\ref{sec:dagger}). Visual encoders remain frozen throughout; backbone updates in Stages~2 and~3 and DAgger use full-parameter fine-tuning.}
\label{tab:training_stages}
\begin{tabular}{@{}>{\raggedright\arraybackslash}p{0.08\linewidth}>{\raggedright\arraybackslash}p{0.22\linewidth}>{\raggedright\arraybackslash}p{0.32\linewidth}>{\raggedright\arraybackslash}p{0.28\linewidth}@{}}
\toprule
Stage & Objective & Training data & Parameter updates \\
\midrule
1 & Vision--language alignment & \texttt{blip\_laion\_cc\_sbu\_558k} alignment collection + human recognition dataset & Adapter only \\
2 & \makecell[l]{Multimodal reasoning\\mid-training} & Stage~1 alignment data + general-purpose VQA data & backbone + adapter \\
3 & Tracking SFT from Stage~2 base & Oracle tracking collection & backbone + adapter + waypoint head \\
\midrule
DAgger & On-policy fine-tuning from Stage~3 & Filtered DAgger rollouts (20\%) + oracle demonstrations (80\%) & backbone + adapter + waypoint head \\
\bottomrule
\end{tabular}
\end{table}

\section{Oracle Data Collection and Supervision}
\label{sec:data}

\paragraph{Privileged oracle and rollout generation.}
We collect training trajectories in Habitat-Sim 0.3.1 using a scripted oracle controller for monocular person following. The oracle uses ground-truth robot and human positions to generate motion commands for following the designated target. It operates in the three \evtbench{} settings introduced with \trackvla{}~\cite{wang2025trackvla}: single-target tracking (\stt{}), distractor tracking (\dtask{}), and ambiguous tracking (\atask{}). These trajectories provide expert behavior-cloning supervision generated from simulator state.

At each environment step, we execute the oracle command and record the egocentric RGB observation, natural-language target description, robot and target positions, and expert action. The action is
\begin{equation}
\vel_t = [v^{\mathrm{fwd}}_t,\,v^{\mathrm{lat}}_t,\,\omega_t],
\end{equation}
where the components denote forward velocity, lateral velocity, and yaw rate in the robot body frame. Observations and state logs are aligned at a 10\,Hz control rate. The learned policy receives RGB history and the target description; privileged simulator state is used to construct supervision.

\paragraph{Training-data scale.}
The oracle collection contains 19,532 episodes and is the sole tracking-data source for Stage~3. Table~\ref{tab:dataset_stats} summarizes its task coverage. Scenes are drawn from HM3D~\cite{ramakrishnan2021hm3d} and Matterport3D~\cite{chang2017matterport3d}. Scene identifiers can appear across task splits, so scene counts are reported separately for each task. Training and benchmark evaluation episodes are disjoint, following the \evtbench{} episode split. The alignment and VQA data used in Stages~1 and~2 are separate from this collection.

\begin{table}[t]
\centering
\caption{Oracle episode collection used for tracking supervision in Stage~3.}
\label{tab:dataset_stats}
\begin{tabular}{lrr}
\toprule
Task & Episodes & Scenes \\
\midrule
\stt{} & 6,906 & 675 \\
\dtask{} & 6,229 & 681 \\
\atask{} & 6,397 & 684 \\
\midrule
Total & \textbf{19,532} & -- \\
\bottomrule
\end{tabular}
\end{table}

\paragraph{Trajectory supervision.}
We convert the recorded robot trajectories into future waypoints in the robot's coordinate frame. Each training window pairs the target description with 31 past RGB observations and the current frame (32 frames total), and provides an eight-step waypoint target with $\Delta t=0.1$\,s. The episode count in Table~\ref{tab:dataset_stats} describes the source collection rather than the number of training windows produced by preprocessing. Stage~3 optimizes the masked waypoint regression objective in Section~\ref{sec:method}. The same oracle also labels policy-visited states during DAgger; rollout collection, label filtering, and the mixture with oracle demonstrations are described in Section~\ref{sec:dagger}.

\section{Experiments}
\label{sec:experiments}

\subsection{Setup}

We evaluate on the \stt{}, \dtask{}, and \atask{} subsets of \evtbench{}~\cite{wang2025trackvla}, with 1,405 episodes per task. \stt{} measures continuous following of a single target. \dtask{} introduces distractors and stresses target identity preservation. \atask{} emphasizes semantic ambiguity among visually similar candidates.

Unless otherwise specified, \method{} uses both frozen visual towers, a 32-frame history, $2\times2$ coarse history pooling, $8\times8$ current-frame pooling, a transformer backbone, and a three-layer waypoint planner. We evaluate both the three-stage model and its DAgger-fine-tuned variant (Section~\ref{sec:training}) on benchmark episodes that are disjoint from training episodes.

Stage~3 uses full-parameter SFT for two epochs; Stages~1 and~2 are summarized by their data, objectives, and trainable modules in Section~\ref{sec:training}.

DAgger fine-tuning initializes from the Stage~3 checkpoint with a fresh optimizer and uses 390,000 samples for one epoch: 20\% filtered DAgger data and 80\% oracle demonstrations, each balanced across tasks (Section~\ref{sec:dagger}). We update the backbone, adapter, and waypoint head with AdamW (weight decay 0.01), a global batch size of 16 on two NVIDIA TITAN RTX GPUs, and bfloat16 mixed precision. The learning rate warms up for 50 steps to $10^{-5}$, then decays with a cosine schedule to $10^{-6}$. Training takes approximately 9.5 hours; we evaluate the final checkpoint.

\subsection{Comparison with Prior Methods}

Table~\ref{tab:main_results} compares both \method{} variants with prior EVT and navigation baselines and the official MiniCPM-RobotTrack model released in OpenBMB's MiniCPM-Robot repository~\cite{openbmbminicpmrobot}. We emphasize SR and CR on DT and AT because successful following among competing candidates and collision avoidance are central to deployment. Higher SR and lower CR are better. STT results for our three-stage and DAgger policies are reported in Table~\ref{tab:dagger_main}. Percentage-point differences below are computed from the displayed two-decimal values.

\begin{table}[t]
\centering
\caption{Comparison on the distractor and ambiguous tracking tasks of \evtbench{}. Scores for IBVS, PoliFormer, EVT, Uni-NaVid, and TrackVLA are taken from TrackVLA~\cite{wang2025trackvla}. Both \method{} models are evaluated on 1,405 episodes per task.}
\label{tab:main_results}
\small
\begin{tabular}{@{}lcccc@{}}
\toprule
 & \multicolumn{2}{c}{\dtask{}} & \multicolumn{2}{c}{\atask{}} \\
\cmidrule(lr){2-3}\cmidrule(lr){4-5}
Method & SR$\uparrow$ & CR$\downarrow$ & SR$\uparrow$ & CR$\downarrow$ \\
\midrule
IBVS$\dagger$~\cite{gupta2016ibvs} & 10.60 & 6.14 & 15.20 & 4.90 \\
PoliFormer$\dagger$~\cite{zeng2025poliformer} & 2.62 & 44.50 & 3.04 & 41.50 \\
EVT~\cite{zhong2024evt} & 3.23 & 47.90 & 17.40 & 45.60 \\
EVT$\ddagger$~\cite{zhong2024evt} & 15.70 & 53.30 & 18.30 & 44.90 \\
Uni-NaVid (Vicuna-7B)~\cite{zhang2025uninavid} & 11.30 & 43.50 & 8.26 & 43.70 \\
TrackVLA (Vicuna-7B)~\cite{wang2025trackvla} & 57.60 & 5.80 & 50.20 & 17.10 \\
MiniCPM-RobotTrack ~\cite{openbmbminicpmrobot} & 53.20 & 13.60 & 58.00 & 9.00 \\
\midrule
Baseline~\cite{omtrackvla2025} & 41.54 & 11.31 & 60.04 & 7.60 \\
\textbf{\method{}} (three-stage) & 55.80 & 12.10 & 60.64 & 9.40 \\
\textbf{\method{}} (three-stage + DAgger) & \textbf{58.22} & 10.11 & \textbf{60.93} & 10.39 \\
\bottomrule
\end{tabular}
\end{table}

The following comparisons first consider the three-stage model. On \dtask{}, it reaches 55.80 SR and 12.10 CR, compared with \trackvla{}'s 57.60 SR and 5.80 CR. Relative to official MiniCPM-RobotTrack, DT success improves by 2.60 percentage points and collision rate decreases by 1.50 points.

On \atask{}, the three-stage model achieves 60.64 SR and 9.40 CR. Relative to \trackvla{}, success increases by 10.44 percentage points and collision rate decreases by 7.70 points. Its success also exceeds MiniCPM-RobotTrack by 2.64 points, with similar CR (9.40 versus 9.00). IBVS has the lowest AT CR at 4.90, illustrating that episode success and collision avoidance capture complementary aspects of following.

Adding DAgger raises DT success to 58.22 and lowers CR to 10.11. The resulting policy has the highest DT and AT success rates in Table~\ref{tab:main_results}, at 58.22 and 60.93, respectively. Its DT success is 0.62 points above \trackvla{} and 5.02 points above MiniCPM-RobotTrack. These comparisons are descriptive; small differences should be interpreted cautiously without repeated-run uncertainty estimates.

Compared with the baseline model (OmTrackVLA)~\cite{omtrackvla2025}, the three-stage policy increases the success of DT from 41.54\% to 55.80\% ($+14.26$ percentage points), while the DT CR increases from 11.31\% to 12.10\%. Subsequent DAgger refinement reaches 58.22\% success and 10.11\% CR, improving both metrics over OmTrackVLA ($+16.68$ SR and $-1.20$ CR points). On AT, the final policy has higher success (60.93\% versus 60.04\%) but also higher CR (10.39\% versus 7.60\%). These results characterize the complete training recipe rather than isolating the effect of alignment alone.

Together, these results position \method{} as a waypoint-prediction baseline with a particularly encouraging profile on ambiguous tracking. End-to-end efficiency also depends on the visual encoders, temporal processing, and deployment hardware, and remains to be measured. The benchmark study characterizes the complete training recipe and architecture.

\subsection{On-Policy Data Aggregation}
\label{sec:dagger}

Behavior cloning trains the policy only on states visited by the expert. At
test time, small errors take the robot into states that are underrepresented in the demonstrations:
the target drifts out of frame, the robot falls behind, or it closes in too
fast. The demonstrations may provide insufficient supervision for recovery from these states. We
address this covariate shift with DAgger~\cite{ross2011dagger}, adapted to a
privileged oracle and a learner conditioned on RGB history and language.

\paragraph{Rollouts and labels.}
The current policy $\pi_i$ drives the robot on training episodes of all three
tasks (\stt{}, \dtask{}, \atask{}), at the 10\,Hz control rate used for the demonstrations.
At every step, the scripted oracle that generated the original demonstrations
computes the command it would issue from the same state. It uses ground-truth
positions of the robot and the humans. The policy's command is executed
($\beta = 0$, where $\beta$ is the probability of executing the oracle action), and the oracle's command is stored only as the supervision label. Frames,
instructions and labels are converted with the same pipeline as the
demonstrations, so DAgger samples and demonstrations share one format.

\paragraph{Filtering privileged labels.}
The oracle is privileged and the learner is not. When the learner has lost
the target, the oracle still issues a confident pursuit command toward a
person who is not in the image. Training on such labels teaches the policy to
accelerate without visual evidence, and in our experiments it raised
collisions sharply. We therefore keep a DAgger sample
only if one of these holds:
\begin{enumerate}
  \item \emph{Framed}: the target is within the success distance and well
        framed in the image ($\mathrm{facing} = 1$).
  \item \emph{Close range}: the robot is closer than 1.2\,m to the target,
        where the oracle backs off. These labels provide supervision for increasing separation.
  \item \emph{Recovery}: the learner had framed the target for at least
        1\,s and then lost it. We keep the steps from the loss until the
        target is framed again, for at most 2\,s.
\end{enumerate}
In recovery states the target is still in the policy's observation history,
providing visual context that may help the policy learn corrective behavior.
These samples capture departures from sustained tracking during the policy's own rollouts.

\paragraph{Aggregation and task balance.}
Following DAgger, each round adds its filtered rollouts to the data of
earlier rounds. We fine-tune from the Stage~3 checkpoint on a fixed budget of
$T = 390{,}000$ samples. DAgger samples make up 20\% of $T$, with an equal number
per task, and samples from the latest policy are used first. The remaining
80\% are demonstrations, again with an equal number per task. We train for
one epoch with learning rate $10^{-5}$, cosine decay, and batch size 16.

\paragraph{Evaluation protocol.}
We evaluate on the full validation sets of \evtbench{} (1{,}405 episodes per
task) and report success rate (SR) and collision rate (CR). The
public evaluation harness initializes the human avatars and the detector's
target identity only on the first episode of each process. Later episodes
can therefore show the wrong people or track the wrong target. We refresh
both on every episode, clear spawn poses left over from the previous
episode, and keep the benchmark's rule that \stt{} episodes contain no
distractors.
In a repeat evaluation of the same model, approximately 23\% of episodes
changed outcome. This observation indicates evaluation variability but does
not by itself establish an uncertainty interval for aggregate SR. We therefore
interpret small SR differences cautiously.

\begin{table}[t]
\centering
\caption{Effect of DAgger, full DT and AT validation sets of \evtbench{} (1{,}405 episodes
per task). }
\label{tab:dagger_main}
\small

\begin{tabular}{lcc cc}
\toprule
 & \multicolumn{2}{c}{\dtask{}} & \multicolumn{2}{c}{\atask{}} \\
\cmidrule(lr){2-3}\cmidrule(lr){4-5}
Model & SR$\uparrow$ & CR$\downarrow$ & SR$\uparrow$ & CR$\downarrow$ \\
\midrule
Oracle (privileged) & 86.74 & 5.17 & 84.92 & 5.00 \\
\midrule
\method{} (three-stage) & 55.80 & 12.10 & 60.64 & 9.40 \\
\method{} (three-stage + DAgger) & \textbf{58.22} & 10.11 & 60.93 & 10.39 \\
\bottomrule
\end{tabular}
\end{table}

\paragraph{Results.}
DAgger improves both success and collision outcomes on \dtask{} (Table~\ref{tab:dagger_main}). Relative to the Stage~3 checkpoint, SR rises from 55.80\% to 58.22\% ($+2.42$ percentage points), while CR falls from 12.10\% to 10.11\% ($-1.99$ points). The collision reduction is concentrated in contacts during forward motion (Section~\ref{sec:collision}). On \atask{}, SR increases slightly from 60.64\% to 60.93\% ($+0.29$ points), while CR increases from 9.40\% to 10.39\% ($+0.99$ points). The observed benefit is therefore concentrated in DT; the small AT success gain does not establish a reliable improvement without repeated-run uncertainty estimates. The privileged oracle achieves 86.74\% DT success and 84.92\% AT success, leaving gaps of 28.52 and 23.99 percentage points, respectively, to the DAgger-refined policy.

\begin{table}[t]
\centering
\small
\caption{Changes from \method{} (three-stage) to \method{} (three-stage + DAgger). Metric differences are percentage points computed from the displayed scores in Table~\ref{tab:dagger_main}; count differences are computed from the episode logs. Positive SR and negative CR indicate improvement.}
\label{tab:dagger_changes}
\begin{tabular}{@{}lrrrr@{}}
\toprule
Task & $\Delta$SR & $\Delta$CR & $\Delta$Successes & $\Delta$Collisions \\
\midrule
DT & +2.42 & -1.99 & +34 & -28 \\
AT & +0.29 & +0.99 & +4 & +14 \\
\bottomrule
\end{tabular}
\end{table}

\section{Collision Analysis}
\label{sec:collision}

Collision avoidance is a major concern for person-following robots in shared spaces. We therefore examine whether the DT success gain after on-policy refinement accompanies fewer contacts. Table~\ref{tab:collision_analysis} summarizes collisions from one evaluation per model on the same 1,405 episode IDs per task. We classify a collision as occurring during forward motion when mean forward velocity over the final 15 logged steps (or all available steps for shorter episodes) is at least 0.15\,m/s. The remaining category includes slow, stopped, and reverse motion; these categories do not establish responsibility for contact.

\begin{table}[t]
\centering
\small
\caption{Collision counts before and after on-policy refinement. Each entry reports three-stage $\rightarrow$ three-stage + DAgger. Forward: mean forward velocity $\geq0.15$\,m/s before contact; below threshold: all remaining collisions.}
\label{tab:collision_analysis}
\begin{tabular}{@{}lccc@{}}
\toprule
Task & All collisions & Forward motion & Below threshold \\
\midrule
DT & $\mathbf{170\rightarrow142}$ & $\mathbf{88\rightarrow56}$ & $82\rightarrow86$ \\
AT & $132\rightarrow146$ & $63\rightarrow68$ & $69\rightarrow78$ \\
\midrule
Overall & $371\rightarrow362$ & $163\rightarrow135$ & $208\rightarrow227$ \\
\bottomrule
\end{tabular}
\end{table}

\paragraph{DT success improves alongside fewer forward-motion collisions.}
On DT, success rises from 55.80\% to 58.22\% while collisions decrease from 170 to 142 (CR 12.10\% to 10.11\%). Forward-motion collisions fall from 88 to 56, including a reduction from 62 to 34 in contacts after step 30. Below-threshold collisions change from 82 to 86. Thus, the observed collision benefit is concentrated in forward motion during distractor tracking. Better use of visual and language context to guide motion is one possible explanation, alongside improved close-range control learned from oracle labels. This comparison does not isolate multimodal understanding as the cause: both variants share the multi-stage recipe, and the logs do not identify which person was hit. Runs are matched by episode ID and instruction, without verified identical simulator random states.

\section{Discussion}
\label{sec:discussion}

\paragraph{Multimodal learning for tracking among distractors.}
The gap between single-target and distractor tracking motivates learning to connect a target description with visual context before optimizing motion. Inspired by staged visual--language learning, \method{} first aligns modalities, then develops multimodal representations, and finally learns waypoint prediction. The increase from 41.54\% DT success with OmTrackVLA~\cite{omtrackvla2025} to 55.80\% with the complete recipe supports this direction. It does not isolate the effect of alignment or individual stages, but is consistent with our hypothesis that multimodal preparation benefits tracking among competing candidates. Temporal context and robot-centric waypoints connect this representation to local control.

\paragraph{Tracking rate and practical following quality.}
Our qualitative real-world observations suggest that high TR alone does not identify the most useful following policy. Effective following also requires remembering where the target moved and using that history to recover after losing sight of it. A robot may need to detour around obstacles, slow down near other people or corners, and temporarily change its following distance. Such behavior can reduce TR while helping the robot continue following without contact. In our deployments, the policy did not always maintain a constant following distance, and lower TR did not necessarily indicate a worse user experience. These observations motivate emphasizing episode success and collision rate in the reported comparisons. They do not establish that lower TR is inherently preferable.

\paragraph{Deployment experience motivates prioritizing DT.}
In our real-world deployments, we qualitatively observed that models with higher DT success provided a better user experience when following a person among distractors. Shared spaces frequently contain other people, making robustness to competing candidates important even when a model already performs well in isolated-person demonstrations. This observation reinforces our emphasis on DT when assessing practical person-following behavior. It is a qualitative deployment observation rather than a controlled user study; the relationship between benchmark DT scores and user experience remains to be quantified.

\paragraph{Promising preliminary on-policy refinement.}
Our DAgger study finds a promising benefit for EVT, particularly DT: supervision on policy-visited states adds 2.42 percentage points of success and reduces collision rate by 1.99 points. The collision analysis associates this gain with fewer contacts during forward motion, suggesting that refinement can improve how the policy acts in distractor-rich situations. Because AT collision rate increases despite a small success-rate gain, the result supports targeted refinement for DT rather than a general improvement across both tasks. The central finding is the potential of combining staged multimodal learning with supervision drawn from the policy's own interactions.

\section{Conclusion}
\label{sec:conclusion}

We presented \method{}, a vision--language policy for short-horizon waypoint prediction in embodied navigation. Frozen DINOv3 and SigLIP features, multiscale temporal context, a transformer backbone, and a lightweight waypoint head connect RGB history and language to robot-centric local trajectories. A three-stage recipe brings together vision--language alignment, multimodal reasoning mid-training, and tracking-specific waypoint supervision, followed by on-policy data aggregation with filtered oracle labels.

Evaluation on \evtbench{} emphasizes distractor and ambiguous tracking, with success rates of 58.22\% and 60.93\% and collision rates of 10.11\% and 10.39\%, respectively. The DT results highlight multi-stage multimodal learning as a promising direction for following a designated person among distractors. Preliminary on-policy refinement further improves DT success and collision outcomes, motivating continued investigation for deployment in shared spaces.

Future work will further investigate DAgger for improving DT, including additional rounds of policy-driven collection, the balance between demonstrations and aggregated rollouts, and the contribution of each label-filtering rule. We will examine whether these refinements improve target retention as well as collision avoidance, and evaluate their transfer to person following in real shared spaces.

\bibliography{example}

\end{document}